# Association rule mining with earthquake data collected from Türkiye region

**Baha ALTURAN[1], İlker TÜRKER[2*]**

[1] Cankiri Karatekin University, Engineering Faculty, Department of Computer Engineering, Cankiri, Türkiye
[2] Karabuk University, Engineering Faculty, Department of Computer Engineering, Karabuk, Türkiye.
iturker@karabuk.edu.tr

**ABSTRACT**

Earthquakes are evaluated among the most destructive disasters for human beings, as also experienced for Türkiye region. Data science has the property of discovering hidden patterns in case a sufficient volume of data is supplied. Time dependency of events, specifically being defined by co-occurrence in a specific time window, may be handled as an associate rule mining task such as a market-basket analysis application. In this regard, we assumed each day's seismic activity as a single basket of events, leading to discovering the association patterns between these events. Consequently, this study presents the most prominent association rules for the earthquakes recorded in Türkiye region in the last 5 years, each year presented separately. Results indicate statistical inference with events recorded from regions of various distances, which could be further verified with geologic evidence from the field. As a result, we believe that the current study may form a statistical basis for the future works with the aid of machine learning algorithm performed for associate rule mining.

**Keywords:** Time series analysis, associate rule mining, machine learning, earthquake dynamics, collective behavior.

## 1. Introduction

Earthquakes, as natural phenomena, pose significant challenges to communities, infrastructures, and disaster management systems worldwide [1]. As our understanding of seismic activities continues to evolve, leveraging advanced technologies becomes imperative for effective earthquake prediction, monitoring, and mitigation [2]. One such technological frontier is the application of data mining techniques, specifically the powerful tool of Association Rule Mining, to seismic data analysis [3].

In recent years, the surge in data generated by seismic sensors, satellite imagery, and geological surveys has opened new avenues for researchers to extract valuable insights into earthquake patterns and behaviors. Associate Rule Mining, a data mining approach initially developed for market basket analysis, has found intriguing applications in uncovering hidden patterns and relationships within complex datasets, including those related to seismic activities [4].

Seismic data collection is a fundamental process in geophysics, integral to the study of the Earth's subsurface structure and composition. The primary method employed for seismic data acquisition involves the controlled generation of seismic waves at the Earth's surface and the subsequent recording of their reflections and refractions by subsurface geological formations. This is typically achieved through the use of seismic sources, such as explosives or specialized vibrators, which emit acoustic waves that penetrate the Earth [5, 6]. These waves interact with the subsurface layers, producing reflections that are detected by an array of sensitive receivers known as geophones or hydrophones strategically placed across the survey area [7]. The recorded data, in the form of seismic traces, are then meticulously processed to enhance signal quality, reduce noise, and transform the information into a coherent image of the subsurface [8]. Processing techniques involve various steps, including time-to-depth conversion, filtering, migration, and velocity analysis. Advanced computational algorithms are applied to correct for complex geological features, ensuring a more accurate representation of subsurface structures. The culmination of these efforts yields detailed seismic images that provide valuable insights into the Earth's interior, aiding in the exploration and characterization of oil and gas reservoirs, geological formations, and other subsurface features [9, 10].

The current study delves into the exploration of Associate Rule Mining as a promising methodology for earthquake data analysis. By examining patterns of co-occurrence and dependencies among various seismic parameters, we aim to unveil hidden associations that can contribute to a deeper comprehension of earthquake dynamics. Such insights have the potential to enhance early warning systems, risk assessments, and overall preparedness for seismic events [11]. In this exploration, we will discuss the foundational principles of Associate Rule Mining and its adaptation to earthquake data. Through the data collected and served by the Kandilli Rasathanesi (Kandilli Observatory and Earthquake Research Institute - KOERI) located in Turkey, we unveil meaningful patterns, providing researchers and seismologists with valuable information to better understand the intricacies of seismic activities in Türkiye region.

As we stand at the intersection of data science and seismology, this article serves as a bridge, connecting the analytical power of Associate Rule Mining with the complex dynamics of earthquake data. By harnessing the potential of these cutting-edge technologies, we aim to contribute to the ongoing efforts to mitigate the impact of earthquakes and bolster our ability to build resilient communities in seismically active regions.

## 2. Data and Methods

### 2.1 Dataset

The seismic data utilized in this study were obtained from KOERI [12]. The dataset spans a period of ~5 years (From the beginning of 2019 to the end of August 2023), capturing seismic events and subsurface information within the Türkiye region with no restriction for magnitude (ML). The seismic recordings were acquired using a network of seismometers strategically positioned to monitor seismic activity across Turkey, made available by a web interface provided by KOERI.

The dataset includes a diverse set of seismic events, ranging from small-scale tremors to larger earthquakes, providing a comprehensive representation of the subsurface dynamics in the study area.

Kandilli Rasathanesi dataset is archived in digital formats, enabling precise analysis and interpretation. has undergone rigorous quality control processes to ensure accuracy and reliability. Each data point in the dataset corresponds to a seismic trace recorded by the seismometer network, capturing the variations in ground motion over time. The seismic data includes information such as event location, depth, magnitude, and waveform characteristics, facilitating a detailed exploration of seismic phenomena. The dataset's accessibility provides a valuable resource for researchers and seismic analysts interested in investigating the geological and seismological aspects of the Turkish region [12].

We retrieved the data from KOERI using the web interface spanning years 2019 to 2023 separately. Since the data for the current year (2023) is made available until the end of October, we used the whole available data for this year.

**2.2. Preprocessing**

We preprocessed the data to filter out seismic events outside the borders of the country, and also having relatively low magnitudes (less than 2.0), reducing the data size gradually as given in Table 1. Due to the diversity of geographic regions with suburban details provided, we processed the regional data as to be unified for city centers, reducing the available number of unique locations. By the way, the resulting data includes only the name of cities and a variety of important locations for earthquake activity such as "*Marmara Denizi*". All preprocessing and computing operations have been performed in Python environment.

**Table 1.** Statistical information about the seismic events within the borders of Türkiye region, spanning the years from 2019 to 2023.

|  | # of all events | # of events within borders | # of events ML ≥ 2.0 within borders | ML (Max) |
|---|---|---|---|---|
| 2019 | 16821 | 13139 | 4948 | 5.7 |
| 2020 | 32230 | 21829 | 8382 | 6.7 |
| 2021 | 23877 | 10717 | 3732 | 5.3 |
| 2022 | 18689 | 14102 | 4268 | 6.1 |
| 2023 (Oct, 31) | 53085 | 48659 | 21491 | 7.5 |

**2.3. Methodology**

We performed associate rule mining with the selection of Apriori algorithm, which is frequently employed for association rule learning tasks. It is used for discovering interesting relationships, associations, or patterns in large datasets. Specifically, Apriori algorithm is employed to identify frequent itemsets in a dataset and generate association rules.

To enable running an association rule mining algorithm, the data needs to be subdivided into virtually generated baskets, based on time dependency. Since association between earthquake events are time-dependent due to the nature of seismic activity influencing each other, a proper time window approach would well suit a definition of baskets collecting subsequent items bought in a single purchase activity. In this regard, we assumed that each single day corresponds to a new time-basket holding associated earthquake events, enabling the usage of Apriori algorithm for mining hidden associations between events. A brief overview of how the Apriori algorithm works is as follows:

*Frequent Itemset Generation:* The algorithm starts by identifying individual items (e.g., products in a store) that have sufficient support in the dataset. Support is a measure of how frequently an item or set of items appears in the dataset.

It then iteratively generates larger itemsets by combining the frequent itemsets discovered in the previous iteration. The process continues until no more frequent itemsets can be generated.

*Association Rule Generation:* Once the frequent itemsets are identified, the algorithm generates association rules based on these itemsets. An association rule typically has the form A → B, where A and B are sets of items.

The confidence of a rule A → B is a measure of how often items in B appear in transactions that contain items in A. High-confidence rules are those where the presence of items in A implies a high likelihood of the presence of items in B.

*Support, Confidence, and Minimum Thresholds:* The Apriori algorithm uses support and confidence thresholds to filter out uninteresting or less significant rules. Users typically set minimum support and confidence values to focus on rules that meet certain criteria.

The Apriori algorithm is named after the "a priori" principle in probability theory, which suggests that if an itemset is frequent, then all of its subsets must also be frequent. This principle helps reduce the search space during the algorithm's execution.

Apriori has been widely used for market basket analysis, where it helps identify associations between products that are frequently purchased together. It has applications in various domains, including retail, healthcare, and web usage mining.

Support, confidence, lift, leverage, and conviction are metrics commonly used in association rule mining to evaluate the strength and significance of discovered patterns. These metrics help assess the quality and reliability of association rules between items in a dataset.

**i. Support:** Support measures the proportion of transactions in a dataset that contain a specific itemset. High support indicates that the itemset is frequently present in the dataset.

$$Support(X) = \frac{Transactions\ containing\ X}{Total\ transactions}$$

**ii. Confidence:** Confidence measures the likelihood that a rule $X \to Y$ holds true, given that $X$ occurs. High confidence implies a strong association between items $X$ and $Y$ when $X$ occurs.

$$Confidence\ (X \to Y) = \frac{Support(X \cup Y)}{Support\ (X)}$$

**iii. Lift:** Lift assesses how much more likely item *Y* is to be purchased when item *X* is purchased, compared to when *Y* is purchased without *X*. Lift > 1 indicates a positive association, suggesting that the occurrence of *X* increases the likelihood of *Y*.

$$Lift\ (X \rightarrow Y) = \frac{Confidence(X \rightarrow Y)}{Support\ (Y)}$$

**iv. Leverage:** Leverage measures the difference between the observed frequency of *X* and *Y* appearing together and the frequency expected if they were independent. A leverage of 0 indicates independence, while positive values suggest a tendency for *X* and *Y* to co-occur.

$$Leverage\ (X \rightarrow Y) = Support(X \cup Y) - Support\ (X) \times Support\ (Y)$$

**v. Conviction:** Conviction assesses the degree to which the consequent *Y* is dependent on the antecedent *X* considering the independence of *X* and *Y*. Higher conviction values indicate stronger dependency between *X* and *Y*.

$$Conviction\ (X \rightarrow Y) = \frac{1 - Support\ (Y)}{1 - Confidence(X \rightarrow Y)}$$

These metrics provide a multifaceted view of association rules, helping analysts and data scientists evaluate the significance and reliability of patterns discovered in transactional datasets.

### 3. Results and Discussion

Having applied Apriori algorithm for the data retrieved from KOERI, we generated the lists of most significant association rules between cities having seismic activity. Though having extracted longer lists, we present only the top 30 rows of highest *confidence* metric, later sorted in decreasing order with respect to the *lift* metric, since this metric mainly handles correlation between the pairs. These associate rules may indicate emergence of geologic correlations between the locations, which should further be evaluated by field experts.

Interpretation of these events should be handled as, a rule from *(Malatya, Manisa)* to *(Elazığ)* means that if earthquakes in *Malatya* and *Elazığ* are recorded in the same day, another event in *Elazığ* has a strong correlation, therefore may be expected.

**Table 2.** Association rules derived from the KOERI dataset, spanning year 2019 for Türkiye region. The rules are listed in descending order wrt. *lift* metric. Top 30 rules are included based on highest *confidence*.

| Antecedent EQ | Consequent EQ | Support | Confidence | Lift | Leverage | Conviction |
| --- | --- | --- | --- | --- | --- | --- |
| (ANKARA) | (DENIZLI) | 0,102 | 0,569 | 1,351 | 0,026 | 1,343 |
| (MARMARA DENIZI) | (MALATYA) | 0,116 | 0,341 | 1,227 | 0,021 | 1,096 |
| (MALATYA) | (MARMARA DENIZI) | 0,116 | 0,416 | 1,227 | 0,021 | 1,132 |
| (DENIZLI) | (DIYARBAKIR) | 0,118 | 0,281 | 1,173 | 0,017 | 1,058 |
| (DIYARBAKIR) | (DENIZLI) | 0,118 | 0,494 | 1,173 | 0,017 | 1,144 |
| (ADANA) | (MUGLA) | 0,102 | 0,685 | 1,162 | 0,014 | 1,304 |
| (KUTAHYA) | (MUGLA) | 0,129 | 0,671 | 1,139 | 0,016 | 1,249 |
| (VAN) | (BALIKESIR) | 0,116 | 0,318 | 1,132 | 0,014 | 1,055 |
| (BALIKESIR) | (VAN) | 0,116 | 0,412 | 1,132 | 0,014 | 1,082 |
| (IZMIR) | (DENIZLI) | 0,105 | 0,469 | 1,113 | 0,011 | 1,090 |
| (SIVAS) | (MUGLA) | 0,105 | 0,655 | 1,111 | 0,010 | 1,190 |
| (MUGLA) | (MALATYA) | 0,182 | 0,308 | 1,108 | 0,018 | 1,044 |
| (MALATYA) | (MUGLA) | 0,182 | 0,653 | 1,108 | 0,018 | 1,184 |
| (KONYA) | (MUGLA) | 0,121 | 0,647 | 1,098 | 0,011 | 1,163 |
| (VAN) | (MARMARA DENIZI) | 0,135 | 0,371 | 1,096 | 0,012 | 1,051 |
| (MARMARA DENIZI) | (VAN) | 0,135 | 0,398 | 1,096 | 0,012 | 1,058 |
| (VAN) | (MALATYA) | 0,110 | 0,303 | 1,089 | 0,009 | 1,036 |
| (MALATYA) | (VAN) | 0,110 | 0,396 | 1,089 | 0,009 | 1,054 |
| (DENIZLI) | (CANAKKALE) | 0,135 | 0,320 | 1,076 | 0,010 | 1,033 |
| (CANAKKALE) | (DENIZLI) | 0,135 | 0,454 | 1,076 | 0,010 | 1,059 |
| (MUGLA) | (DIYARBAKIR) | 0,149 | 0,252 | 1,053 | 0,007 | 1,017 |
| (DIYARBAKIR) | (MUGLA) | 0,149 | 0,621 | 1,053 | 0,007 | 1,082 |
| (MUGLA) | (CANAKKALE) | 0,185 | 0,313 | 1,052 | 0,009 | 1,023 |
| (CANAKKALE) | (MUGLA) | 0,185 | 0,620 | 1,052 | 0,009 | 1,081 |
| (MANISA) | (VAN) | 0,143 | 0,377 | 1,036 | 0,005 | 1,021 |
| (VAN) | (MANISA) | 0,143 | 0,394 | 1,036 | 0,005 | 1,023 |
| (MUGLA) | (MARMARA DENIZI) | 0,207 | 0,350 | 1,034 | 0,007 | 1,018 |
| (MARMARA DENIZI) | (MUGLA) | 0,207 | 0,610 | 1,034 | 0,007 | 1,052 |
| (DENIZLI) | (MALATYA) | 0,121 | 0,288 | 1,034 | 0,004 | 1,013 |
| (MALATYA) | (DENIZLI) | 0,121 | 0,436 | 1,034 | 0,004 | 1,025 |

Results for year 2019 given in Table 2 indicate that regions *Ankara* and *Denizli* have the highest correlation with one of the highest confidence level, where *Malatya* and *Marmara Denizi* have the highest correlation in two directions. *Denizli* and *Diyarbakır* exhibits the same behavior though being very distant geographically. This type of associations between distant cities are evident for other pairs such as *Adana* and *Muğla*, *Van* and *Balıkesir* (in two directions), *Sivas* and *Muğla* etc., which may be caused by frequent seismic activity in one of the locations (i.e. *Denizli, Muğla*), therefore qualifying for a statistical inference. Therefore, geologic correspondence needs further consideration supported by physical evidence. The most prominent seismic activity in year 2019 was in *Denizli* with magnitude 6.0, recorded on August 8[th] [13]. Therefore, this region has remarkable weight of occurrence in Table 2, while the following active region *Marmara Denizi* (Earthquake of *Silivri* with magnitude 5.7) has comparable association instances.

**Table 3.** Association rules derived from the KOERI dataset, spanning year 2020 for Türkiye region. The rules are listed in descending order wrt. *lift* metric. Top 30 rules are included based on highest *confidence*.

| Antecedent EQ | Consequent EQ | Support | Confidence | Lift | Leverage | Conviction |
|---|---|---|---|---|---|---|
| (ELAZIG, VAN) | (MALATYA, MANISA) | 0,184 | 0,615 | 1,253 | 0,037 | 1,323 |
| (VAN, MANISA) | (ELAZIG, MALATYA) | 0,184 | 0,598 | 1,241 | 0,036 | 1,289 |
| (MALATYA, VAN) | (ELAZIG, MANISA) | 0,184 | 0,615 | 1,240 | 0,035 | 1,308 |
| (MALATYA, MUGLA) | (ELAZIG, MANISA) | 0,156 | 0,613 | 1,236 | 0,030 | 1,302 |
| (MALATYA, VAN, MANISA) | (ELAZIG) | 0,184 | 0,827 | 1,232 | 0,035 | 1,902 |
| (MALATYA, ELAZIG, VAN) | (MANISA) | 0,184 | 0,827 | 1,198 | 0,030 | 1,791 |
| (MALATYA, MUGLA, MANISA) | (ELAZIG) | 0,156 | 0,792 | 1,179 | 0,024 | 1,578 |
| (ELAZIG, MALATYA, MUGLA) | (MANISA) | 0,156 | 0,814 | 1,179 | 0,024 | 1,667 |
| (ELAZIG, MUGLA) | (MALATYA, MANISA) | 0,156 | 0,576 | 1,174 | 0,023 | 1,201 |
| (MARMARA DENIZI, ELAZIG) | (MALATYA) | 0,162 | 0,787 | 1,172 | 0,024 | 1,541 |
| (VAN, MANISA) | (ELAZIG) | 0,241 | 0,786 | 1,171 | 0,035 | 1,534 |
| (ELAZIG, VAN) | (MANISA) | 0,241 | 0,807 | 1,169 | 0,035 | 1,607 |
| (MUGLA, MANISA) | (ELAZIG, MALATYA) | 0,156 | 0,559 | 1,159 | 0,021 | 1,174 |
| (MARMARA DENIZI, MANISA) | (MALATYA) | 0,167 | 0,772 | 1,150 | 0,022 | 1,443 |
| (ELAZIG, VAN, MANISA) | (MALATYA) | 0,184 | 0,761 | 1,134 | 0,022 | 1,378 |
| (ELAZIG, MALATYA, MANISA) | (VAN) | 0,184 | 0,500 | 1,134 | 0,022 | 1,118 |
| (VAN) | (ELAZIG, MALATYA, MANISA) | 0,184 | 0,416 | 1,134 | 0,022 | 1,084 |
| (DENIZLI) | (MANISA) | 0,153 | 0,778 | 1,127 | 0,017 | 1,393 |
| (KUTAHYA) | (MALATYA) | 0,184 | 0,753 | 1,122 | 0,020 | 1,330 |
| (MALATYA, MUGLA) | (ELAZIG) | 0,192 | 0,753 | 1,121 | 0,021 | 1,329 |
| (MALATYA, MUGLA) | (MANISA) | 0,197 | 0,774 | 1,121 | 0,021 | 1,371 |
| (ELAZIG, MUGLA, MANISA) | (MALATYA) | 0,156 | 0,750 | 1,117 | 0,016 | 1,315 |
| (MALATYA, MANISA) | (ELAZIG) | 0,367 | 0,749 | 1,115 | 0,038 | 1,308 |
| (ELAZIG) | (MALATYA, MANISA) | 0,367 | 0,547 | 1,115 | 0,038 | 1,125 |
| (ELAZIG, MUGLA) | (MANISA) | 0,208 | 0,768 | 1,112 | 0,021 | 1,333 |
| (MUGLA, MANISA) | (ELAZIG) | 0,208 | 0,745 | 1,110 | 0,021 | 1,290 |
| (MALATYA, VAN) | (ELAZIG) | 0,222 | 0,743 | 1,107 | 0,021 | 1,280 |
| (ELAZIG, VAN) | (MALATYA) | 0,222 | 0,743 | 1,107 | 0,021 | 1,280 |
| (MARMARA DENIZI, MALATYA) | (MANISA) | 0,167 | 0,763 | 1,104 | 0,016 | 1,304 |
| (ELAZIG, MANISA) | (MALATYA) | 0,367 | 0,740 | 1,103 | 0,034 | 1,266 |

2020 was an active year for Türkiye region with prominent earthquakes recorded in *Sivrice, Elazığ*, closely following the earthquake of lower magnitude (5.6 and later 5.5) recorded for *Akhisar, Manisa* by 2 days. This consequence may also be an indicator of time dependency between seismic events despite being distant geographically. *Karlıova, Bingöl* (5.9), *Marmaris, Muğla* (5.4) and *Pütürge, Malatya* (5.7) were the other active regions for the rest of the year [13]. The association rules generated from 2020 records is given in Table 3, prominently featuring correlations between *Elazığ* and *Manisa* regions, also include instances from closer locations such as *Malatya* and *Van*. Occurrence of earthquakes in *Marmara Denizi* in correlation with these points of seismic activity is also engaging, with some examples such as [(*Marmara Denizi, Elazığ*) to (*Malatya*)] or, [(*Marmara Denizi, Manisa*) to (*Malatya*)].

**Table 4.** Association rules derived from the KOERI dataset, spanning year 2021 for Türkiye region. The rules are listed in descending order wrt. *lift* metric. Top 30 rules are included based on highest *confidence*.

| Antecedent EQ | Consequent EQ | Support | Confidence | Lift | Leverage | Conviction |
|---|---|---|---|---|---|---|
| (BALIKESIR, ELAZIG) | (MANISA) | 0,077 | 0,509 | 1,471 | 0,025 | 1,332 |
| (KUTAHYA) | (MALATYA) | 0,088 | 0,516 | 1,456 | 0,028 | 1,334 |
| (ELAZIG, MALATYA) | (MANISA) | 0,071 | 0,491 | 1,417 | 0,021 | 1,283 |
| (BALIKESIR, MANISA) | (ELAZIG) | 0,077 | 0,609 | 1,411 | 0,022 | 1,453 |
| (SIVAS) | (MALATYA) | 0,096 | 0,479 | 1,353 | 0,025 | 1,240 |
| (ELAZIG, MANISA) | (BALIKESIR) | 0,077 | 0,438 | 1,338 | 0,019 | 1,197 |
| (ERZURUM) | (MALATYA) | 0,091 | 0,458 | 1,293 | 0,021 | 1,192 |
| (ANKARA) | (MUGLA) | 0,080 | 0,414 | 1,278 | 0,017 | 1,154 |
| (ERZINCAN) | (BALIKESIR) | 0,080 | 0,414 | 1,267 | 0,017 | 1,149 |
| (IZMIR) | (MANISA) | 0,096 | 0,427 | 1,233 | 0,018 | 1,141 |
| (ERZURUM) | (BALIKESIR) | 0,080 | 0,403 | 1,232 | 0,015 | 1,127 |
| (MALATYA, MANISA) | (ELAZIG) | 0,071 | 0,531 | 1,230 | 0,013 | 1,212 |
| (ANTALYA) | (VAN) | 0,096 | 0,407 | 1,224 | 0,018 | 1,126 |
| (IZMIR) | (ELAZIG) | 0,118 | 0,524 | 1,216 | 0,021 | 1,196 |
| (IZMIR) | (MALATYA) | 0,096 | 0,427 | 1,204 | 0,016 | 1,126 |
| (ERZURUM) | (MANISA) | 0,082 | 0,417 | 1,204 | 0,014 | 1,121 |
| (SIVAS) | (MANISA) | 0,082 | 0,411 | 1,187 | 0,013 | 1,110 |
| (MANISA) | (ELAZIG) | 0,176 | 0,508 | 1,178 | 0,027 | 1,156 |
| (ELAZIG) | (MANISA) | 0,176 | 0,408 | 1,178 | 0,027 | 1,104 |
| (ERZURUM) | (VAN) | 0,077 | 0,389 | 1,170 | 0,011 | 1,092 |
| (ERZINCAN) | (MALATYA) | 0,080 | 0,414 | 1,169 | 0,012 | 1,102 |
| (BALIKESIR) | (MALATYA) | 0,135 | 0,412 | 1,162 | 0,019 | 1,098 |
| (MALATYA) | (BALIKESIR) | 0,135 | 0,380 | 1,162 | 0,019 | 1,085 |
| (ERZINCAN) | (MANISA) | 0,077 | 0,400 | 1,156 | 0,010 | 1,090 |
| (SIVAS) | (VAN) | 0,077 | 0,384 | 1,154 | 0,010 | 1,083 |
| (ANTALYA) | (MUGLA) | 0,088 | 0,372 | 1,148 | 0,011 | 1,076 |
| (ELAZIG, MANISA) | (MALATYA) | 0,071 | 0,406 | 1,146 | 0,009 | 1,087 |
| (SIVAS) | (ELAZIG) | 0,099 | 0,493 | 1,143 | 0,012 | 1,122 |
| (KONYA) | (MALATYA) | 0,080 | 0,397 | 1,121 | 0,009 | 1,071 |
| (MANISA) | (BALIKESIR) | 0,126 | 0,365 | 1,117 | 0,013 | 1,060 |

In contrast with the previous year, 2021 does not yield significant seismic activity with no moderate or high impact events. However, the association rules presented in Table 4 indicates the continuum of earthquakes with relatively lower magnitudes as a consequence of the events of the previous year. Despite being low in magnitude, correlations of the events are very high as traced in the values of lift coefficient. *Balıkesir*, *Elazığ* and *Manisa* seem to be prominently correlated, while *Malatya* region seems to be affected by the countrywide earthquakes of lower magnitude.

*Gölyaka*, *Düzce* region experienced the strongest earthquake of 2022 with magnitude 6.0, following the *Buca*, *İzmir* earthquake (4.5) that had happened in the same month, *November*. (*Kütahya*) to (*Van*, *Muğla*) is the strongest association rule for the year while *Adana* to *Muğla*, *Kahramanmaraş* to *Elazığ*, *Marmara Denizi* to *Manisa* are other noteworthy associations, given in Table 5.

**Table 5.** Association rules derived from the KOERI dataset, spanning year 2022 for Türkiye region. The rules are listed in descending order wrt. *lift* metric. Top 30 rules are included based on highest *confidence*.

| Antecedent EQ | Consequent EQ | Support | Confidence | Lift | Leverage | Conviction |
| --- | --- | --- | --- | --- | --- | --- |
| (KUTAHYA) | (VAN, MUGLA) | 0,082 | 0,375 | 1,468 | 0,026 | 1,191 |
| (ADANA) | (MUGLA) | 0,091 | 0,805 | 1,332 | 0,023 | 2,027 |
| (MUGLA, KUTAHYA) | (VAN) | 0,082 | 0,556 | 1,305 | 0,019 | 1,292 |
| (KAHRAMANMARAS) | (ELAZIG) | 0,080 | 0,460 | 1,269 | 0,017 | 1,181 |
| (MARMARA DENIZI) | (MANISA) | 0,074 | 0,351 | 1,264 | 0,015 | 1,113 |
| (VAN, KUTAHYA) | (MUGLA) | 0,082 | 0,750 | 1,241 | 0,016 | 1,582 |
| (ANTALYA, VAN) | (MUGLA) | 0,071 | 0,743 | 1,229 | 0,013 | 1,538 |
| (CANAKKALE) | (ELAZIG) | 0,074 | 0,443 | 1,221 | 0,013 | 1,144 |
| (CORUM) | (MUGLA) | 0,082 | 0,732 | 1,211 | 0,014 | 1,475 |
| (KUTAHYA) | (VAN) | 0,110 | 0,500 | 1,174 | 0,016 | 1,148 |
| (MUGLA, MANISA) | (ELAZIG) | 0,074 | 0,422 | 1,163 | 0,010 | 1,102 |
| (ELAZIG, MANISA) | (MUGLA) | 0,074 | 0,692 | 1,145 | 0,009 | 1,286 |
| (KUTAHYA) | (ELAZIG) | 0,091 | 0,413 | 1,138 | 0,011 | 1,085 |
| (MUGLA, MANISA) | (VAN) | 0,085 | 0,484 | 1,138 | 0,010 | 1,114 |
| (KUTAHYA) | (MUGLA) | 0,148 | 0,675 | 1,117 | 0,016 | 1,217 |
| (MARMARA DENIZI) | (ELAZIG) | 0,085 | 0,403 | 1,110 | 0,008 | 1,067 |
| (MALATYA) | (ELAZIG) | 0,110 | 0,400 | 1,103 | 0,010 | 1,062 |
| (BALIKESIR, ELAZIG) | (MUGLA) | 0,071 | 0,667 | 1,103 | 0,007 | 1,187 |
| (BALIKESIR, MUGLA) | (VAN) | 0,085 | 0,470 | 1,103 | 0,008 | 1,083 |
| (MALATYA, VAN) | (MUGLA) | 0,077 | 0,667 | 1,103 | 0,007 | 1,187 |
| (DENIZLI) | (VAN) | 0,102 | 0,468 | 1,100 | 0,009 | 1,080 |
| (IZMIR) | (ANKARA) | 0,077 | 0,359 | 1,098 | 0,007 | 1,050 |
| (MARMARA DENIZI) | (MUGLA) | 0,140 | 0,662 | 1,096 | 0,012 | 1,172 |
| (MANISA) | (VAN) | 0,129 | 0,465 | 1,093 | 0,011 | 1,074 |
| (VAN, MANISA) | (MUGLA) | 0,085 | 0,660 | 1,091 | 0,007 | 1,162 |
| (ANTALYA) | (ELAZIG) | 0,099 | 0,396 | 1,091 | 0,008 | 1,055 |
| (ANTALYA, MUGLA) | (VAN) | 0,071 | 0,464 | 1,090 | 0,006 | 1,072 |
| (ELAZIG) | (ANKARA) | 0,129 | 0,356 | 1,089 | 0,011 | 1,045 |
| (ANKARA) | (ELAZIG) | 0,129 | 0,395 | 1,089 | 0,011 | 1,053 |
| (BALIKESIR, MUGLA) | (ELAZIG) | 0,071 | 0,394 | 1,086 | 0,006 | 1,052 |

Another noteworthy outcome from the previous tables is that active regions such as *Elazığ*, *Malatya* and *Van* continuously influence the strong associations for all years spanned in this study. This activity would result in the devastating earthquakes of 2023 recorded with thousands of events with various scale of magnitudes. Association rules for the same year are presented in Table 6 with the domination of rules generated by the mentioned regions. The whole table holds correlations between the cities affected by the disaster, with no prominent direction or specification among them.

**Table 6.** Association rules derived from the KOERI dataset, spanning year 2023 for Türkiye region. The rules are listed in descending order wrt. *lift* metric. Top 30 rules are included based on highest *confidence*.

| Antecedent EQ | Consequent EQ | Support | Confidence | Lift | Leverage | Conviction |
|---|---|---|---|---|---|---|
| (HATAY, ADANA, K.MARAS) | (ADIYAMAN, MALATYA) | 0,530 | 0,843 | 1,244 | 0,104 | 2,052 |
| (ADIYAMAN, MALATYA) | (HATAY, ADANA, K.MARAS) | 0,530 | 0,782 | 1,244 | 0,104 | 1,702 |
| (HATAY, ADANA) | (ADIYAMAN, MALATYA) | 0,530 | 0,839 | 1,237 | 0,102 | 1,997 |
| (HATAY, ADANA) | (ADIYAMAN, MALATYA, K.MARAS) | 0,530 | 0,839 | 1,237 | 0,102 | 1,997 |
| (ADIYAMAN, MALATYA) | (HATAY, ADANA) | 0,530 | 0,782 | 1,237 | 0,102 | 1,687 |
| (ADIYAMAN, MALATYA, K.MARAS) | (HATAY, ADANA) | 0,530 | 0,782 | 1,237 | 0,102 | 1,687 |
| (HATAY, ADANA, MALATYA) | (ADIYAMAN, K.MARAS) | 0,530 | 0,843 | 1,232 | 0,100 | 2,011 |
| (ADIYAMAN, K.MARAS) | (HATAY, ADANA, MALATYA) | 0,530 | 0,774 | 1,232 | 0,100 | 1,645 |
| (ADIYAMAN, HATAY, MALATYA) | (ADANA, K.MARAS) | 0,530 | 0,875 | 1,231 | 0,100 | 2,316 |
| (ADANA, K.MARAS) | (ADIYAMAN, HATAY, MALATYA) | 0,530 | 0,745 | 1,231 | 0,100 | 1,550 |
| (HATAY, ADANA, K.MARAS) | (ADIYAMAN) | 0,530 | 0,843 | 1,226 | 0,098 | 1,990 |
| (HATAY, ADANA, MALATYA) | (ADIYAMAN) | 0,530 | 0,843 | 1,226 | 0,098 | 1,990 |
| (MALATYA, HATAY, ADANA, K.MARAS) | (ADIYAMAN) | 0,530 | 0,843 | 1,226 | 0,098 | 1,990 |
| (ADIYAMAN) | (HATAY, ADANA, K.MARAS) | 0,530 | 0,770 | 1,226 | 0,098 | 1,618 |
| (ADIYAMAN) | (HATAY, ADANA, MALATYA) | 0,530 | 0,770 | 1,226 | 0,098 | 1,618 |
| (ADIYAMAN) | (MALATYA, HATAY, ADANA, K.MARAS) | 0,530 | 0,770 | 1,226 | 0,098 | 1,618 |
| (HATAY, ADANA) | (ADIYAMAN, K.MARAS) | 0,530 | 0,839 | 1,226 | 0,097 | 1,956 |
| (ADIYAMAN, K.MARAS) | (HATAY, ADANA) | 0,530 | 0,774 | 1,226 | 0,097 | 1,630 |
| (HATAY, ADANA) | (ADIYAMAN) | 0,530 | 0,839 | 1,220 | 0,095 | 1,935 |
| (ADIYAMAN) | (HATAY, ADANA) | 0,530 | 0,770 | 1,220 | 0,095 | 1,604 |
| (HATAY, ADIYAMAN) | (ADANA, K.MARAS) | 0,530 | 0,866 | 1,218 | 0,095 | 2,154 |
| (HATAY, ADIYAMAN) | (MALATYA, ADANA, K.MARAS) | 0,530 | 0,866 | 1,218 | 0,095 | 2,154 |
| (ADANA, K.MARAS) | (HATAY, ADIYAMAN) | 0,530 | 0,745 | 1,218 | 0,095 | 1,524 |
| (MALATYA, ADANA, K.MARAS) | (HATAY, ADIYAMAN) | 0,530 | 0,745 | 1,218 | 0,095 | 1,524 |
| (ADIYAMAN, MALATYA) | (ADANA, K.MARAS) | 0,586 | 0,864 | 1,216 | 0,104 | 2,130 |
| (ADANA, K.MARAS) | (ADIYAMAN, MALATYA) | 0,586 | 0,824 | 1,216 | 0,104 | 1,832 |
| (MALATYA, ELAZIG, K.MARAS) | (ADIYAMAN) | 0,520 | 0,836 | 1,216 | 0,092 | 1,905 |
| (ADIYAMAN) | (MALATYA, ELAZIG, K.MARAS) | 0,520 | 0,756 | 1,216 | 0,092 | 1,550 |
| (ELAZIG, MALATYA) | (ADIYAMAN, K.MARAS) | 0,520 | 0,832 | 1,215 | 0,092 | 1,875 |
| (ADIYAMAN, K.MARAS) | (ELAZIG, MALATYA) | 0,520 | 0,760 | 1,215 | 0,092 | 1,560 |

## 4. Conclusions

The current study evaluates the seismic events as a collection of time-related events, handling each day's earthquakes as a separate basket of events. By the way, a procedure similar to market-basket analysis is performed for these events, outlining the relations between a variety of close to distant earthquakes. These outcomes may be an indicator of geologic correspondence between these events even if they are significantly distant. Geologic validation can be recommended as a complementary work for the current study.